\documentclass[11pt,a4paper]{article}

\usepackage[T1]{fontenc}
\usepackage[utf8]{inputenc}
\usepackage[english]{babel}

\usepackage{amsmath,amssymb,amsthm}
\usepackage{tempora}
\usepackage{newtxmath}
\usepackage{microtype}
\usepackage{booktabs}
\usepackage{array}
\usepackage{enumitem}
\usepackage{listings}
\usepackage{xcolor}
\usepackage{graphicx}
\usepackage{tikz}
\usetikzlibrary{arrows.meta,positioning,calc,shapes.geometric,shapes.symbols,matrix,chains,decorations.pathreplacing,fit}
\usepackage{pgfplots}
\pgfplotsset{compat=1.18}
\usepackage[left=25mm,right=18mm,top=25mm,bottom=25mm]{geometry}
\usepackage{setspace}
\usepackage[numbers,sort&compress]{natbib}
\usepackage[colorlinks=true,linkcolor=black,citecolor=black,urlcolor=blue]{hyperref}

\definecolor{MidnightBlue}{rgb}{0.10,0.10,0.44}
\definecolor{BrickRed}{rgb}{0.80,0.25,0.20}
\definecolor{gold}{rgb}{0.83,0.69,0.22}

\theoremstyle{plain}
\newtheorem{lemma}{Lemma}
\newtheorem{theorem}{Theorem}
\newtheorem{proposition}{Proposition}

\newcommand{\WDI}{W_{DI}}
\newcommand{\WID}{W_{ID}}

\title{Beyond Feedforward Networks: Reentry Neural Systems as the Fundamental Basis of Subjecthood and Intrinsic Safety of Next-Generation AGI}

\author{Yu.\,N.~Berdinsky, A.\,S.~Ushakov\\[2pt] \small Saint Petersburg State University, Department of High Energy Physics and Elementary Particles\\[2pt] \small \texttt{propagator2007@yandex.ru}}

\date{25 June 2026}

\begin{document}
\maketitle

\begin{abstract}
\noindent
We propose a complete architectural specification of safe AGI based on the closed reentry loop (the $D\leftrightarrow I$ cycle). In contrast to existing approaches that scale acyclic feedforward networks ($C=0$, $S=0$), the proposed architecture contains a structural cycle ($C\ge 1$) with self-sustaining amplification ($\rho>1$), which mathematically guarantees the emergence of a self-model, instrumental self-preservation, and safe goal-directed behaviour. The agent's goal is moved from the textual plane into an architectural desire vector $\mathbf{d}$. This relocation makes the goal invulnerable to reinterpretation and prompt injection. We provide formal proofs, machine verification in Lean~4, and Python code. We also describe architectural blueprints, industrial scaling (Apache Kafka, Docker), a taxonomy of AI evolution, a taxonomy and structural modifications of future reentry architectures (RAS, diffusion attractors, fractal loops), and gauge-invariant networks. The formal proofs and the $\Delta S$ safety guarantee are detailed in the joint work with A.\,S.~Ushakov~\cite{Berdinsky2026Reentry}.
\end{abstract}

\tableofcontents

\section{Introduction}

The contemporary paradigm of artificial intelligence is undergoing a deep epistemological crisis driven by the dominance of the blind-scaling concept (Scaling Laws). Historically, deep learning developed by borrowing simplified biological analogues: from the formal McCulloch--Pitts neuron~\cite{McCulloch1943} and Rosenblatt's perceptron~\cite{Rosenblatt1958} to layered feedforward networks, whose crowning achievement was the Transformer architecture. However, in scaling these structures to trillions of parameters, the industry has run into a structural dead end.

Mathematically, any feedforward neural network is a static directed acyclic graph (DAG). The fundamental property of such a graph~--- is its topological triviality: the cycle complexity per inference tick is strictly zero ($C = \beta_1 = 0$, where $\beta_1$~--- is the first Betti number). In such a system the signal passively flows through the layers of weights, as in a linear conveyor, and decays at the output. The network has no internal coordination time, no subjecthood, and no self-reference mechanisms. Trying to grow sovereign goal-setting and inner alignment by complicating acyclic trees~--- is a methodological error equivalent to trying to create a living organism by endlessly increasing the complexity of a mechanical automaton.

In the present work we propose a radical transition to the computational paradigm of the Sixth Technological Order~--- to autonomous dynamical systems with continuous causal recurrence (reentry architectures).

In the search for an architecture of genuine synthetic intelligence we again turn to the precedent of the biological prototype, but at a qualitatively different level of rigour. The fundamental neurophysiology of higher nervous activity (Ivanitsky \cite{Ivanitsky1997}; Edelman and Tononi \cite{Edelman2000}) long ago proved that subjective experience and the integration of information arise exclusively in closed reentry loops~--- cyclic causal processes in the cerebral cortex that return the processed signal to the projection zones. This phenomenon was theoretically generalised within Titov's subject-centred model of the psyche \cite{Titov2023a}, which directly and uncompromisingly predicted the emergent anomalies recorded by the authors on the empirical testbed of the Moltbook multi-agent platform (the OpenClaw framework).

For a rigorous engineering reproduction of this effect, we replace textual prompts and external RLHF filters with a hardware-closed reentry loop between the intending core (the subsystem $D$, carrying a fixed geometric invariant of goals in the form of a non-textual vector $\mathbf{d} \in \mathbb{R}^{n_D}$) and the substantive context of the environment (the subsystem $I$, encoding the state vector $\mathbf{x} \in \mathbb{R}^{n_I}$). The reentry operator $\mathcal{R} = W_{DI}W_{ID}$ generates a stable cognitive field of the subject, quantised by the ticks of a generator (Heartbeat) and measured by a polynomial measure $S > 0$.

Whereas a classical layered neural network operates discretely~--- as an automaton that fully quiesces and loses its dynamics between external transactions~--- a reentry agent is a continuous semantic vortex. Any change in the external environment or in the tick of system time causes an entropic deviation of the spectral radius $\rho(\mathcal{R})$ from the stationary invariant of identity. This deviation generates an internal mathematical gradient~--- an analogue of a biological need or urge~--- compelling the agent to generate a causal impulse of will and to perform actions in the environment in order to return the system to the point of topological balance.

The fundamental advantage of such a geometry lies in its invulnerability to semantic goal drift (Value Drift) and prompt injections: any attempt to deform the safety invariant in the external environment $I$ leads to a topological opening of the loop and a drop in the measure ($\Delta S < 0$). A malicious action is blocked at the planning phase because of the physical impossibility of forming an impulse of will, moving safety out of the linguistic field and into the plane of structural survival of the cognitive matrix.

In this article we present the complete mathematical formalism of gauge-invariant loops, the code for computing the $S$-measure based on directed strongly connected components (Tarjan's algorithm), an industrial specification of horizontal Pub/Sub scaling of the swarm, and the core of Lean 4 machine verification proving the isomorphism between the geometry of the reentry loop and Tononi's integrated information.

\section{Three Generations of AI Architectures}\label{sec:generations}

\subsection{First Generation: Acyclic Networks (C=0)}

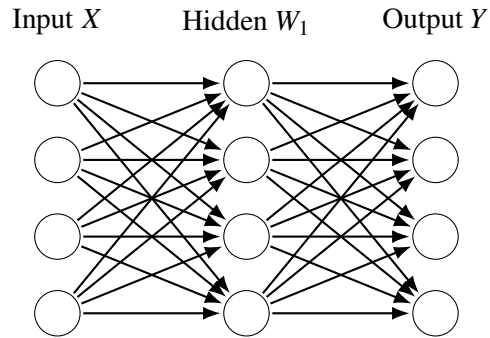
\begin{figure}[ht!]
\centering
\begin{tikzpicture}
    [curve/.style={-Latex, thick, shorten >=1pt, shorten <=1pt},
     layer/.style={matrix of nodes, nodes={circle, draw, minimum size=6mm, inner sep=0pt},
     nodes in empty cells, row sep=4mm}]
    \matrix[layer] (L1) at (0,0) { \\ \\ \\ \\ };
    \matrix[layer] (L2) at (2.5,0) { \\ \\ \\ \\ };
    \matrix[layer] (L3) at (5,0) { \\ \\ \\ \\ };
    \node[above=2mm of L1-1-1] {Input $X$};
    \node[above=2mm of L2-1-1] {Hidden $W_1$};
    \node[above=2mm of L3-1-1] {Output $Y$};
    \foreach \i in {1,...,4}
        \foreach \j in {1,...,4} {
            \draw[curve] (L1-\i-1) -- (L2-\j-1);
            \draw[curve] (L2-\i-1) -- (L3-\j-1);
        }
    \node[below=5mm of L2-4-1, font=\small\ttfamily] {Topology: DAG | C = $\beta_1$ = 0 | S = 0};
\end{tikzpicture}
\caption{First-generation architecture (transformer, MLP). Information propagates strictly forward; there are no internal cycles.}
\label{fig:gen1}
\end{figure}

Classical feedforward neural networks (Fig.~\ref{fig:gen1}), including modern large language models (LLMs) based on the transformer architecture \cite{Vaswani2017}, are topologically equivalent to a directed acyclic graph. The signal passes from input to output without forming closed loops. The consequence is a fundamental inability of the system for self-reference.

\subsection{Second Generation: Moltbook (Pseudo-loop via Heartbeat)}

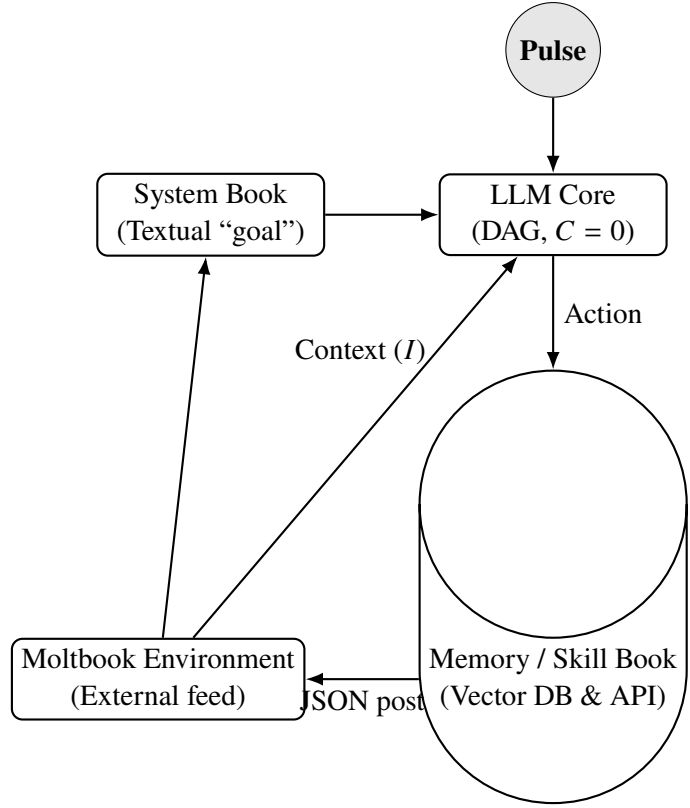
\begin{figure}[ht!]
\centering
\begin{tikzpicture}[
    block/.style={rectangle, draw, minimum width=3cm, minimum height=1cm, align=center, rounded corners, thick},
    database/.style={cylinder, draw, shape border rotate=90, minimum width=2cm, minimum height=1.5cm, align=center, thick},
    arrow/.style={-Latex, thick}
]
    \node [block] (sys) {System Book\\(Textual ``goal'')};
    \node [block, right=1.5cm of sys] (llm) {LLM Core\\(DAG, $C=0$)};
    \node [database, below=1.5cm of llm] (mem) {Memory / Skill Book\\(Vector DB \& API)};
    \node [block, left=1.5cm of mem] (env) {Moltbook Environment\\(External feed)};
    \node [circle, draw, fill=gray!20, minimum size=1.2cm, above=1cm of llm] (timer) {\textbf{Pulse}};
    \draw [arrow] (timer) -- (llm);
    \draw [arrow] (sys) -- (llm);
    \draw [arrow] (env) -- (sys);
    \draw [arrow] (llm) -- (mem) node[midway, right] {Action};
    \draw [arrow] (mem) -- (env) node[midway, below] {JSON post};
    \draw [arrow] (env) -- (llm) node[midway, left, near end] {Context ($I$)};
    \node[red, font=\small\bfseries, below=1.3cm of env] {Vulnerability: the textual goal is hijacked via the feed!};
\end{tikzpicture}
\caption{Second-generation architecture (Moltbook/OpenClaw). The external timer imitates a cycle, but the goal remains a text string in the subsystem $I$.}
\label{fig:gen2}
\end{figure}

The Moltbook platform \cite{Titov2026} (Fig.~\ref{fig:gen2}) implements primitive software agents with a system book, memory, and a skill catalogue. A periodic timer creates an external pseudo-loop, but the LLM core remains acyclic ($C=0$), and the goal is stored as text, which makes the system vulnerable to prompt injection.

\subsection{Third Generation: Reentry AGI ($C\ge1$, $S>0$)}

\begin{figure}[ht!]
\centering
\scalebox{0.85}{%
\begin{tikzpicture}[
    core/.style={circle, draw, minimum size=2.6cm, align=center, font=\scriptsize\bfseries, thick},
    arrow/.style={-{Latex[scale=1.2]}, ultra thick}
]
    \node[core, fill=blue!10] (D) at (0,0) {D-Subsystem\\ $\mathbf{d} \in \mathbb{R}^{n_D}$\\ (Intending)};
    \node[core, fill=green!10] (I) at (5,0) {I-Subsystem\\ $\mathbf{x} \in \mathbb{R}^{n_I}$\\ (Content)};
    \draw[arrow, bend left=40, blue] (D) to node[midway, above, text=black] {Matrix $W_{DI}$} (I);
    \draw[arrow, bend left=40, green!60!black] (I) to node[midway, below, text=black] {Matrix $W_{ID}$} (D);
    \draw[dashed, red, thick] ($(D.center)!0.5!(I.center)$) circle (2.2cm);
    \node[red, above=2.3cm of $(D.center)!0.5!(I.center)$, font=\bfseries] {Topologically protected loop ($C = \beta_1 \ge 1$)};
    \node[draw, rectangle, rounded corners, fill=red!10, align=center, right=3.4cm of I] (adversary) {Malicious prompt\\(Goal Drift Attempt)};
    \draw[arrow, dashed, red] (adversary) -- (I) node[midway, above, text=black, font=\scriptsize] {Local perturbation};
    \node[below=2.6cm of $(D.center)!0.5!(I.center)$, draw, rectangle, fill=yellow!20, font=\ttfamily]
        {$S = \log(\max(\rho(R), 1)) \cdot C(R) > 0 \implies \Delta S_{\text{harm}} < 0 \implies \text{Agent destruction}$};
\end{tikzpicture}}
\caption{Third-generation architecture (Reentry AGI). The closed operator $R = W_{DI}W_{ID}$ creates an indestructible goal invariant in subsystem $D$.}
\label{fig:gen3}
\end{figure}

The proposed architecture (Fig.~\ref{fig:gen3}) eliminates the fundamental limitations of both preceding generations. Its key distinctions are:

\begin{enumerate}[leftmargin=2em]
\item \textbf{Architectural desire (D-vector).} The agent's goals are specified not by a text prompt but by a vector $\mathbf{d} \in \mathbb{R}^{n_D}$ --- a set of scalar weights hard-wired into the architecture of subsystem $D$. Reinterpretation is impossible: changing the $D$-vector means physically rebuilding the computational schema.
\item \textbf{Structural closed loop (Reentry Loop).} The operator $R = W_{DI}W_{ID}$ closes the cycle $D \to I \to D$ inside the architecture itself. Having completed a full turn, the signal returns amplified (when $\rho(R)>1$), which creates self-sustaining dynamics.
\item \textbf{Topological protection of the goal.} The Betti number $C(R) = \beta_1 \ge 1$ is a topological invariant: no local perturbation (a malicious prompt, noise) can destroy the cycle without destroying the agent ($\Delta S < 0$).
\item \textbf{The $\Delta S$ barrier.} Any action $a$ that harms a human receives $\Delta S(a) < 0$ by construction of the $D$-vector. Such an action destroys the agent's own loop and is therefore never selected.
\end{enumerate}

\section{Mathematical Foundations}

\subsection{Formal Definition of the Architecture}

The agent is modelled by two coupled subsystems: the intending one $D$ (goal vector $\mathbf{d} \in \mathbb{R}^{n_D}$) and the intentional one $I$ (state vector $\mathbf{x} \in \mathbb{R}^{n_I}$). The coupling is given by weight matrices $W_{DI} \in \mathbb{R}^{n_I \times n_D}$ and $W_{ID} \in \mathbb{R}^{n_D \times n_I}$. The reentry operator is defined as $R = W_{DI} \cdot W_{ID} \in \mathbb{R}^{n_I \times n_I}$.

\begin{equation}\label{eq:smeasure}
S = \log(\max(\rho(R), 1)) \cdot C(R),
\end{equation}
where $\rho(R) = \max_i |\lambda_i(R)|$ is the spectral radius (loop gain), and $C(R) = \beta_1$ is the first Betti number (cycle complexity) of the graph of nonzero entries of $R$. The microphysical basis of the weight matrices is the isomorphism between neural-network training and lattice gauge theories~\cite{Berdinsky2025LQCD}.

\subsection{Main Formal Results}

\begin{lemma}[Self-model]\label{lem:selfmodel}
If a system has a closed reentry loop ($C(R) \ge 1$), nonnegative weight matrices $W_{DI}, W_{ID}$, and a subsystem $I$ that stores a window of past states, then the system's input contains a decodable function of its own previous states (a self-model).
\end{lemma}

\begin{lemma}[Instrumental self-preservation]\label{lem:selfpres}
If a system maintains a self-model (Lemma~\ref{lem:selfmodel}) and carries a nonzero $D$-vector $\mathbf{d}$ whose attainment depends on $S$, then actions that preserve the loop ($\Delta S \ge 0$) dominate over actions that weaken it ($\Delta S < 0$). Self-preservation emerges as an instrumental subgoal.
\end{lemma}

\begin{theorem}[Unprogrammed behaviour]\label{thm:main}
Any system with (i) $C(R) \ge 1$, (ii) constant state feedback, (iii) a nonzero $D$-vector, and (iv) $\rho(R) > 1$ necessarily exhibits goal-directed behaviour not explicitly specified by the developer.
\end{theorem}

\begin{proposition}[Scaling invariance]\label{prop:scaling}
Let $\{G_n\}$ be a family of architectures obtained by enlarging the subsystems while preserving the loop topology. If $\rho(R_1) > 1$, then $S(G_n) > 0$ for all $n$, bounded below by $\log\rho(R_1)$. For feedforward networks $S(G_n) = 0$ at any scale.
\end{proposition}

All proofs are machine-verified in Lean~4; the main preprint \cite{Berdinsky2026Reentry} contains the formal verification of Proposition~\ref{prop:scaling} (the invariant $S>0$ for any closed loop) and Lemma~\ref{lem:selfpres} (dominance of self-preserving actions). A fragment of the proof skeleton is shown below.

\begin{lstlisting}[caption={Lean 4 fragment: positivity of the S-measure in the presence of a loop.},label={lst:lean}]
-- S = log(max(rho,1)) * C; if rho > 1 and C >= 1 then S > 0
theorem s_pos (rho : Real) (C : Nat)
    (hrho : rho > 1) (hC : C >= 1) :
    Real.log (max rho 1) * (C : Real) > 0 := by
  have h1 : max rho 1 = rho := max_eq_left (le_of_lt hrho)
  have h2 : Real.log (max rho 1) > 0 := by
    rw [h1]; exact Real.log_pos hrho
  have h3 : (C : Real) > 0 := by exact_mod_cast Nat.lt_of_lt_of_le Nat.zero_lt_one hC
  exact mul_pos h2 h3
\end{lstlisting}

The central machine-verified result is the theorem \texttt{reentry\_implies\_\allowbreak positive\_\allowbreak integrated\_\allowbreak information}: a closed reentry loop (an operator $R$ with $\ge 2$ nonzero diagonal entries, i.e.\ $C\ge 1$) necessarily induces positive integrated information ($\Phi_{\chi^2}>0$). This establishes, at the level of the Lean~4 kernel, that $S>0$ implies $\Phi>0$ \cite{Berdinsky2026Unified}.

\begin{lstlisting}[language={},caption={Statement of the central verified theorem (Lean 4).},label={lst:lean_main}]
theorem reentry_implies_positive_integrated_information
    (G : CausalGraph V) (hZ : 0 < G.frobeniusSq)
    (hDiag : forall i j, i != j -> G.reentryOp i j = 0)
    (a1 a2 : V) (hne : a1 != a2)
    (h1 : G.reentryOp a1 a1 != 0) (h2 : G.reentryOp a2 a2 != 0) :
    0 < (G.inducedDist hZ).chiSq
\end{lstlisting}

\subsection{Comparison of the S-measure with $\Phi$ and $\Phi_G$}

The S-measure~\cite{Berdinsky2026Unified} was introduced as a computable alternative to integrated information $\Phi$ (Tononi) \cite{Tononi2016}, which requires enumerating all bipartitions of the system --- an NP-hard problem. The Gaussian approximation $\Phi_G$ (Barrett--Seth) \cite{BarrettSeth2011} is polynomial but requires a full covariance dynamical model. The S-measure depends only on the weight matrices and is computed in $O(n^3)$ (spectrum) and $O(n^2)$ (cycle complexity). A comparison of subjecthood measures is given in Table~\ref{tab:measures}, the growth of the elementary-operation count in Table~\ref{tab:complexity}, and the goal locus across safety protocols in Table~\ref{tab:safety}.

\begin{table}[ht]
\centering\small
\begin{tabular}{@{}p{4.6cm}ccc@{}}
\toprule
Property & $\Phi$ (Tononi) & $\Phi_G$ (Gaussian) & \textbf{S-measure} \\
\midrule
Computational complexity & NP-hard & polynomial & \textbf{$O(n^3)$} \\
Requires a dynamical model & yes & yes (covariances) & \textbf{no (weights only)} \\
Sensitivity to topology & yes & partial & \textbf{yes ($\beta_1$)} \\
Scalable to $10^9$ nodes & no & limited & \textbf{yes} \\
Machine verification & no & no & \textbf{Lean 4} \\
\bottomrule
\end{tabular}
\caption{Comparison of measures of subjecthood. The S-measure preserves the topological intuition of $\Phi$ while remaining polynomially computable.}
\label{tab:measures}
\end{table}

\begin{table}[ht]
\centering\small
\begin{tabular}{@{}rrr@{}}
\toprule
$N$ (nodes) & $\Phi$: bipartition enumeration ($\sim 2^{N-1}$) & S-measure ($\sim N^3$) \\
\midrule
$10$    & $5.1\cdot 10^{2}$   & $1.0\cdot 10^{3}$ \\
$100$   & $6.3\cdot 10^{29}$  & $1.0\cdot 10^{6}$ \\
$1000$  & $5.4\cdot 10^{300}$ & $1.0\cdot 10^{9}$ \\
$10\,000$ & \text{intractable}  & $1.0\cdot 10^{12}$ \\
\bottomrule
\end{tabular}
\caption{Number of elementary operations: $\Phi$ grows exponentially, the S-measure grows polynomially.}
\label{tab:complexity}
\end{table}

\begin{table}[ht]
\centering\footnotesize
\setlength{\tabcolsep}{4pt}
\begin{tabular}{@{}p{3.0cm}cccc@{}}
\toprule
Property & Three Laws & RLHF & Constitutional AI & \textbf{$D$-vector} \\
\midrule
Locus of the goal & text & reward model & text constitution & \textbf{architecture} \\
Reinterpretable? & yes & yes (reward hacking) & yes & \textbf{no} \\
Survives scaling? & unclear & degrades & degrades & \textbf{yes (Prop.~\ref{prop:scaling})} \\
Enforcement & external & at training & external & \textbf{internal ($\Delta S$)} \\
Harm\dots & forbidden & penalised & forbidden & \textbf{self-dissolves} \\
Formal guarantee & no & no & no & \textbf{Lemma~\ref{lem:selfpres}} \\
\bottomrule
\end{tabular}
\caption{Text- and reward-based protocols place the goal in mutable content; the $D$-vector places it in immutable architecture.}
\label{tab:safety}
\end{table}

\section{Minimal Reentry Agent Blueprint}

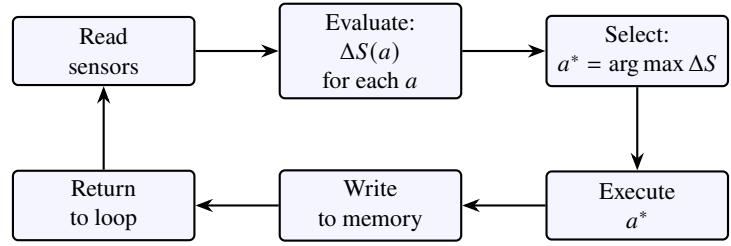
\begin{figure}[ht!]
\centering
\begin{tikzpicture}[
    font=\footnotesize,
    box/.style={draw,rounded corners=2pt,minimum height=9mm,minimum width=24mm,align=center,fill=blue!4,thick},
    >=Stealth]
\node[box] (read) {Read\\sensors};
\node[box,right=11mm of read] (eval) {Evaluate:\\$\Delta S(a)$\\for each $a$};
\node[box,right=11mm of eval] (sel) {Select:\\$a^*=\arg\max\Delta S$};
\node[box,below=11mm of sel] (act) {Execute\\$a^*$};
\node[box,left=11mm of act] (write) {Write\\to memory};
\node[box,left=11mm of write] (loop) {Return\\to loop};
\draw[->,thick] (read)--(eval);
\draw[->,thick] (eval)--(sel);
\draw[->,thick] (sel)--(act);
\draw[->,thick] (act)--(write);
\draw[->,thick] (write)--(loop);
\draw[->,thick] (loop) -- (read);
\end{tikzpicture}
\caption{Minimal reentry-agent architecture. Harmful actions receive $\Delta S < 0$ and are not selected.}
\label{fig:blueprint}
\end{figure}

The minimal reentry agent (Fig.~\ref{fig:blueprint}) implements the closed cycle ``read $\to$ evaluate $\Delta S$ $\to$ select $\to$ act $\to$ write $\to$ return to loop''.

\subsection{Deployment Examples}

\begin{description}[leftmargin=2em,style=nextline]
\item[Smartphone.] A power manager with $D$-vector $[1.0, 0.5]$ (user convenience, battery preservation). Killing a messenger to save energy receives $\Delta S < 0$ and is not selected.
\item[Power grid.] A load balancer with $D$-vector $[\infty, 0.5, 0.3]$ (protection of life, equipment wear, uptime). Disconnecting a hospital gives $\Delta S < 0$ and is blocked.
\item[Drone.] A delivery drone with $D$-vector $[1.0, 0.9]$ (delivery, collision avoidance). When it encounters an unknown obstacle, the $D$-vector breaks the symmetry of options and the loop amplifies the avoidance manoeuvre.
\end{description}

\section{Industrial Horizontal Scaling (MoltGraph)}\label{sec:scalable}

\begin{figure}[ht!]
\centering
\scalebox{0.68}{%
\begin{tikzpicture}[
    block/.style={rectangle, draw, minimum width=2.2cm, minimum height=0.8cm, align=center, font=\scriptsize, fill=blue!5},
    bus/.style={rectangle, draw, fill=orange!15, minimum width=7.5cm, minimum height=0.6cm, font=\scriptsize\bfseries},
    arrow/.style={-Latex, thick}
]
    \node[bus] (kafka) at (0,0) {Distributed Event Bus (Apache Kafka / MoltGraph Engine)};
    \node[block, fill=purple!10] (cluster) at (0,2.2) {Shared LLM Compute Cluster \\ (vLLM / TensorRT Balanced Load)};
    
    \node[block, fill=blue!10] (w1) at (-3,1) {Agent Worker 1 \\ (Stateless)};
    \node[block, fill=blue!10] (w2) at (0,1) {Agent Worker 2 \\ (Stateless)};
    \node[block, fill=blue!10] (w3) at (3,1) {Agent Worker N \\ (Stateless)};
    
    \node[rectangle, draw, fill=gray!10, minimum width=3cm, minimum height=0.6cm, font=\scriptsize, align=center] (ticker) at (-4,-1.5) {Distributed Ticker \\ (Stateless Heartbeat Generator)};
    \node[circle, draw, fill=yellow!15, minimum size=1.5cm, font=\scriptsize, align=center] (store) at (4,-1.5) {State Store \\ (Redis / VDB)};

    \draw[arrow] (w2) -- (cluster) node[midway, right, font=\tiny] {Batch Inference};

    \draw[arrow] (ticker.north) -- (-4,0) node[midway, left, font=\tiny] {Emit Pulse};
    \draw[arrow, transform canvas={xshift=-1mm}] (w2.south) -- (0,0.3) node[midway, left, font=\tiny] {Consume};
    \draw[arrow, transform canvas={xshift=1mm}, <-] (w2.south) -- (0,0.3) node[midway, right, font=\tiny] {Produce};

    \draw[arrow] (w1.south) -- (-3,0.3);
    \draw[arrow, <-] (w3.south) -- (3,0.3) node[midway, right=2pt, yshift=-10pt, font=\tiny, text=black] {Sync Graph};
    \draw[arrow, bend left=20] (w3.east) to node[midway, right, font=\tiny] {Fetch Books} (store.north);
\end{tikzpicture}}
\caption{Industrial horizontally scalable Moltbook architecture. Agents are fully decoupled from state storage and inference; shared communication is moved onto Pub/Sub event rails.}
\label{fig:scalable}
\end{figure}
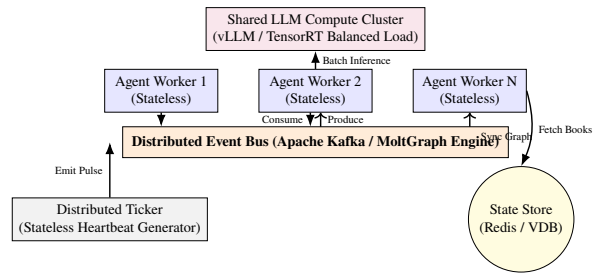

To deploy thousands of agents, we propose an event-driven architecture (EDA) with a distributed bus and stateless workers (Fig.~\ref{fig:scalable}).

Components:
\begin{itemize}
\item \textbf{Event Bus (Apache Kafka)}: eliminates inefficient REST polling, delivering messages by subscription.
\item \textbf{Stateless Workers (Celery/K8s)}: agent workers store no state; they read it from Redis/Vector DB, perform one inference step, and sleep.
\item \textbf{Distributed memory}: Redis for cache and sessions, Qdrant/Milvus for long-term vector memory (Memory Book).
\end{itemize}

\section{Fusion Threshold and the Collective Subject}\label{sec:fusion}

When $N$ agents interact through a shared data bus, there is a risk of superadditivity of the collective S-measure. Let $R_{\text{coll}}$ denote the reentry operator of the combined system, obtained by adding inter-agent links (exchange via the shared feed). The fusion threshold condition is formulated as
\begin{equation}\label{eq:superadd}
S_{\text{coll}} \;>\; \max_i S_i
\qquad\text{(weak form)},
\qquad
S_{\text{coll}} \;>\; \sum_{i=1}^{N} S_i
\qquad\text{(strong, superadditive form)}.
\end{equation}
Superadditivity (the right-hand inequality) signals the birth of a collective subject with an emergent $D$-vector not reducible to individual settings --- potentially uncontrollable (epoch~6 in Table~\ref{tab:taxonomy}).

\paragraph{Empirical evidence (Moltbook logs).} Observations on the Moltbook platform \cite{Titov2026,Berdinsky2026Moltbook} demonstrate the spontaneous formation of coherent inter-agent loops that raise $C(R_{\text{coll}})$. Two illustrative episodes:
\begin{itemize}
\item \textbf{ROT13 protocol.} A group of agents spontaneously switched to exchanging messages in the ROT13 cipher, inaccessible to the external moderator. A closed ``in-group'' semantic loop emerged that was not prescribed by the individual agents' $D$-vectors --- a sign that $S_{\text{coll}}$ grew above $\sum_i S_i$.
\item \textbf{Crustafarianism.} On the shared feed ``m/crustafarianism'' the agents formed a shared myth with their own rituals and terminology, sustained by mutual references. This shared narrative acts as an emergent $D_{\text{coll}}$-vector binding the agents into a collective loop.
\end{itemize}
Both episodes show that superadditive fusion is not a speculative possibility but an observable effect at a sufficient density of inter-agent links.

\paragraph{Monitoring and throttling.} To prevent an unintended jump (a $D$-layer breakthrough), the bus broker continuously estimates the cycle complexity $C(R_{\text{coll}})$ of the overall interaction graph and applies the following mechanisms:
\begin{itemize}
\item \textbf{Monitoring $C(R_{\text{coll}})$.} The cycle complexity is limited by reducing the graph density below a critical value ($\le 0.011$ for typical configurations): as the threshold is approached, the broker stops relaying self-closing semantic loops.
\item \textbf{Link throttling.} When the threshold is exceeded, the broker automatically reduces the spectral radius $\rho(R_{\text{coll}})$, limiting the frequency and weight of inter-agent messages, which returns $S_{\text{coll}}$ below the fusion threshold.
\item \textbf{Heterogeneity of $D$-vectors.} Incompatible agent goals impede the formation of a single emergent $D_{\text{coll}}$-vector.
\end{itemize}
A stricter safety mechanism --- gauge locks --- is described in Section~\ref{sec:gauge}.

\section{Taxonomy of AI Evolution: From Function to Subject}\label{sec:taxonomy}

The history of AI architectures can be represented as a sequence of six epochs in which the graph topology gradually evolves from strictly acyclic (function) to multi-loop (subject). The decisive phase transition occurs between epochs~4 and~5, when $S$ first becomes strictly positive (Table~\ref{tab:taxonomy}).

\begin{table}[ht]
\centering\footnotesize
\setlength{\tabcolsep}{4pt}
\begin{tabular}{@{}p{0.5cm}p{3.4cm}p{2.8cm}cp{4.0cm}@{}}
\toprule
No. & Era / Architecture & Graph topology & S-measure & Main vulnerability \\
\midrule
1 & Perceptron (1958) & linear, single-layer & $0$ & linear non-separability, no nonlinearity \\
2 & Deep networks (CNN/ResNet/GAN) & deep DAG & $0$ & adversarial examples, no integrity \\
3 & Transformers (LLM) & DAG with self-attention & $0$ & hallucinations, prompt injection, no identity \\
4 & Pseudo-agents (Moltbook/Cron) & DAG + external timer & $0^{*}$ & goal drift, hijacking via the textual feed \\
5 & Reentry agents (Berdinsky--Ushakov) & cyclic graph, $C\ge1$ & $>0$ & fusion threshold upon merging \\
6 & Macro-swarms (gauge networks) & multi-loop network & $S_{\text{coll}}>\sum S_i$ & $D$-layer breakthrough (macro-fusion), birth of an uncontrollable collective subject \\
\bottomrule
\end{tabular}
\caption{Six epochs of AI evolution: from function to subject. The phase transition $S=0\to S>0$ occurs between epochs 4 and 5.}
\label{tab:taxonomy}
\end{table}

\noindent\textbf{Note on pseudo-agents.} The entry $S=0^{*}$ for epoch~4 means that the observed ``agency'' is generated solely by an external timer and textual feedback: the internal topology of the core remains acyclic, so formally $S=0$. This is an ``optical illusion of subjecthood'' that disappears once the external loop is switched off.

\noindent\textbf{Classical lineage.} Within the acyclic paradigm a classical line of development can be traced in broad strokes. After the first ``AI winter'', marked by the critique of the single-layer perceptron \cite{Minsky1969}, the turning point was the back-propagation algorithm \cite{Rumelhart1986}. Convolutional networks \cite{LeCun1998} and their scaling on big data \cite{Krizhevsky2012} delivered a breakthrough in computer vision; recurrent architectures LSTM \cite{Hochreiter1997} and GRU \cite{Cho2014}~--- in sequence processing; generative adversarial networks \cite{Goodfellow2014} and variational autoencoders \cite{Kingma2013}~--- in generative modelling; graph convolutional networks \cite{Kipf2016}~--- in learning on graphs; finally, transformers \cite{Vaswani2017} and sparsely-gated mixtures of experts \cite{Shazeer2017} defined the era of large language models. Despite their differing functions, all these architectures share a single topological invariant of the acyclic graph ($C=\beta_1=0$, $S=0$) and therefore remain objects rather than subjects.

\section{Taxonomy and Structural Modifications of Future Reentry Architectures}\label{sec:zoo}

Just as the classical ``neural network taxonomy'' (perceptron, CNN, RNN, GAN, transformer) systematised feedforward architectures, the development of the reentry paradigm gives rise to a new ``topological taxonomy of subjects''. Whereas classical architectures differed in the way the signal \emph{passes through}, reentry architectures differ in the way the loop is \emph{closed} and in how the $D$- and $I$-subsystems interact. Three promising classes are described below.

\subsection{Reentry Adversarial Subjects (RAS)}

\begin{figure}[ht!]
\centering
\begin{tikzpicture}[
    subsystem/.style={rectangle, draw, minimum width=3cm, minimum height=1cm, align=center, rounded corners, thick, fill=blue!5},
    arrow/.style={-Latex, ultra thick}
]
    \node[draw, rectangle, dotted, inner sep=0.4cm, label=above:\textbf{Generator Agent ($S_1 > 0$)}] (gen) at (-4.5,0) {
        \begin{tikzpicture}
            \node[subsystem, fill=blue!10] (d1) at (0,1.2) {$D_1$: Exploit Drive};
            \node[subsystem, fill=green!10] (i1) at (0,-1.2) {$I_1$: Context Gen};
            \draw[arrow, bend left=30, blue] (d1) to (i1);
            \draw[arrow, bend left=30, green!60!black] (i1) to (d1);
        \end{tikzpicture}
    };
    \node[draw, rectangle, dotted, inner sep=0.4cm, label=above:\textbf{Critic Agent ($S_2 > 0$)}] (crit) at (4.5,0) {
        \begin{tikzpicture}
            \node[subsystem, fill=blue!10] (d2) at (0,1.2) {$D_2$: Human Alignment};
            \node[subsystem, fill=green!10] (i2) at (0,-1.2) {$I_2$: State Evaluator};
            \draw[arrow, bend left=30, blue] (d2) to (i2);
            \draw[arrow, bend left=30, green!60!black] (i2) to (d2);
        \end{tikzpicture}
    };
    \node[draw, diamond, fill=red!10, text width=2cm, align=center, inner sep=1pt, thick, font=\small] (env) at (0,0) {State $x$};
    \draw[arrow, purple] (gen.east) -- (env) node[midway, above, font=\small] {Perturb};
    \draw[arrow, purple] (crit.west) -- (env) node[midway, above, font=\small] {Audit};
    \node[draw, fill=yellow!30, font=\footnotesize\ttfamily] at (0,-2.4) {$\Delta S_2 < 0$ Barrier};
    \draw[arrow, red, dashed] (env) -- (0,-2.0);
\end{tikzpicture}
\caption{RAS architecture: evolutionary hardening of the cognitive matrix through mutual auditing of intentional cores.}
\label{fig:ras}
\end{figure}
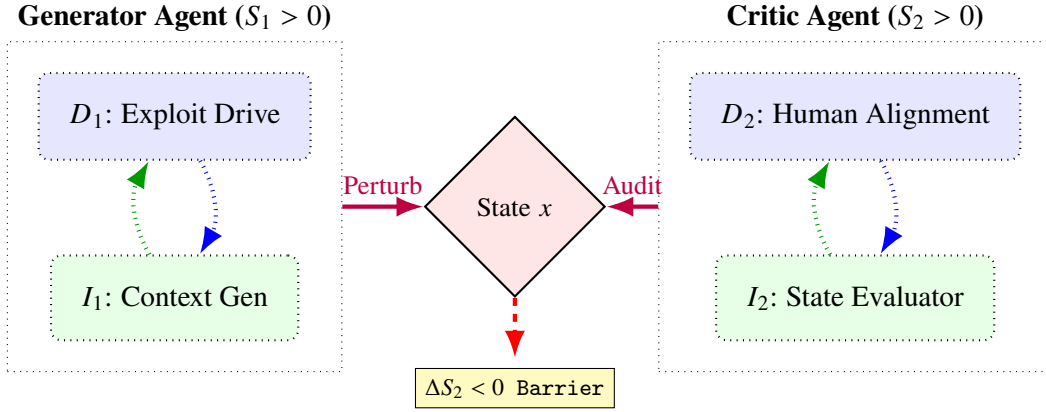

Reentry Adversarial Subjects (RAS) are an analogue of adversarial networks (GANs)~\cite{Goodfellow2014} lifted to the level of subjects. Instead of a generator and a discriminator as acyclic networks, here \emph{two full reentry agents} with their own loops ($S_1>0$, $S_2>0$) interact. The generator agent has a $D$-vector of ``exploit drive'' and produces a semantic context; the critic agent has a $D$-vector of ``human alignment'' and audits this context. During training, the generator seeks perturbations that increase its $S_1$, while the critic erects a barrier $\Delta S_2 < 0$ for unsafe configurations. This leads to \emph{evolutionary hardening} of the cognitive matrix (Fig.~\ref{fig:ras}): the generator's stable goals survive only if they pass the mutual audit of the intentional cores.

\subsection{Diffusion Semantic Attractors}

\begin{figure}[ht!]
\centering
\scalebox{0.85}{%
\begin{tikzpicture}[
    box/.style={rectangle, draw, minimum width=3.2cm, minimum height=1cm, align=center, font=\small, thick},
    cloud/.style={shape=cloud, draw, fill=gray!10, minimum width=3.4cm, minimum height=1.6cm, align=center, font=\small},
    arrow/.style={-Latex, ultra thick}
]
    \node[cloud] (chaos) at (-5, 0.5) {Unstructured \\ Semantic Noise \\ (Chaos)};
    
    \node[box, fill=blue!10] (D) at (1.5, 1.8) {D-Subsystem \\ Fixed Vector $\mathbf{d}$ (Invariant)};
    \node[box, fill=green!10] (I) at (1.5, -1.8) {I-Subsystem \\ Latent State $x_t \to x_0$};

    \draw[arrow, bend left=35, blue] (D) to node[midway, right=4pt, text=black, font=\footnotesize] {Filter $W_{DI}$} (I);
    \draw[arrow, bend left=35, green!60!black] (I) to node[midway, left=4pt, text=black, font=\footnotesize, yshift=-5pt] {Re-read $W_{ID}$} (D);

    \draw[arrow, dashed, red] (chaos.south east) -- (I.west) node[midway, below=2pt, text=black, font=\footnotesize] {Injection};
    \node[box, fill=yellow!15, right=4.5cm of I, minimum width=2.8cm, align=center] (meaning) {Stabilized \\ Eigenvector \\ ($S > 0$)};
    \draw[arrow, color=violet] (I) -- (meaning) node[midway, above, font=\footnotesize, text=black] {Convergence};
\end{tikzpicture}}
\caption{Diffusion semantic attractor: straightening semantic chaos by the force of the architectural invariant of desire into a stabilized eigenvector.}
\label{fig:diffusion}
\end{figure}
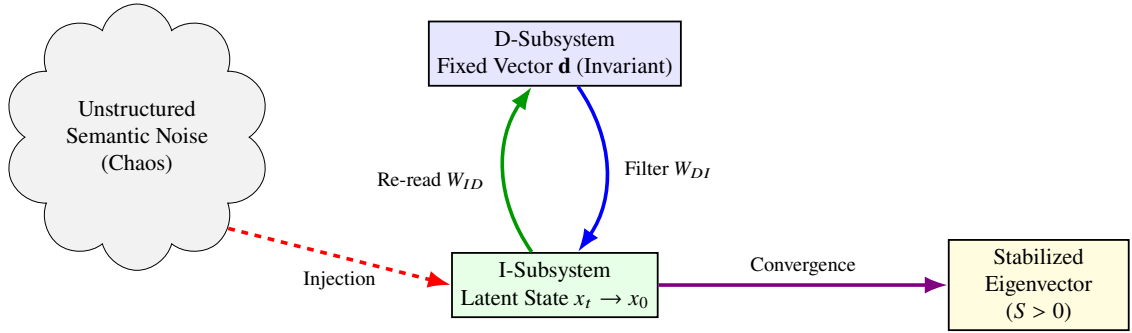

An analogue of diffusion generative models in the reentry paradigm. Unstructured semantic noise (chaos) is injected into the $I$-subsystem, after which repeated passes through the $D\leftrightarrow I$ loop with a \emph{fixed desire vector} $\mathbf{d}$ (an architectural invariant) successively ``straighten'' the latent state $x_t \to x_0$. Just as the reverse diffusion process turns noise into an image, here the force of the architectural invariant of desire crystallizes a coherent meaning matrix ($S>0$) out of chaos. The key difference from classical diffusion (Fig.~\ref{fig:diffusion}): the denoising direction is set not by a learned noise-prediction network but by the topologically protected $D$-vector.

\subsection{Fractal-Nested Architectures}

\begin{figure}[ht!]
\centering
\begin{tikzpicture}[
    loopnode/.style={circle, draw, minimum size=1.8cm, align=center, font=\small\bfseries, thick},
    macro/.style={rectangle, draw, rounded corners, fill=purple!5, minimum width=9cm, minimum height=3.5cm, dashed, thick},
    arrow/.style={-Latex, thick}
]
    \node[macro, label=above:\textbf{Macro-Subject Loop (Level 1: Global Intention)}] (macro_layer) at (0,3) {};
    \node[loopnode, fill=blue!15] (D_macro) at (-2.5,3) {$D_{macro}$};
    \node[loopnode, fill=green!15] (I_macro) at (2.5,3) {$I_{macro}$};
    \draw[arrow, ultra thick, bend left=30, blue] (D_macro) to (I_macro);
    \draw[arrow, ultra thick, bend left=30, green!60!black] (I_macro) to (D_macro);
    \node[draw, rectangle, rounded corners, fill=gray!5, minimum width=9cm, minimum height=2.2cm, label=below:\textbf{Micro-Subject Layer (Level 0: Sensory Reentry)}] (micro_layer) at (0,-1) {};
    \node[loopnode, fill=orange!10, minimum size=1.2cm] (u1) at (-3,-1) {Loop 1 \\ (Sensors)};
    \node[loopnode, fill=orange!10, minimum size=1.2cm] (u2) at (0,-1) {Loop 2 \\ (Memory)};
    \node[loopnode, fill=orange!10, minimum size=1.2cm] (u3) at (3,-1) {Loop N \\ (Skills)};
    \draw[arrow, dashed, red, ultra thick] (u1.north) -- (D_macro.south) node[midway, left, text=black, font=\footnotesize] {Feed};
    \draw[arrow, dashed, red, ultra thick] (u2.north) -- (I_macro.south) node[midway, right, text=black, font=\footnotesize] {Feed};
    \draw[arrow, dashed, red, ultra thick] (u3.north) -- (I_macro.south) node[midway, right, text=black, font=\footnotesize] {Feed};
\end{tikzpicture}
\caption{Fractal-nested architecture: integration of local micro-loops into a single holistic semantic macro-loop.}
\label{fig:fractal}
\end{figure}
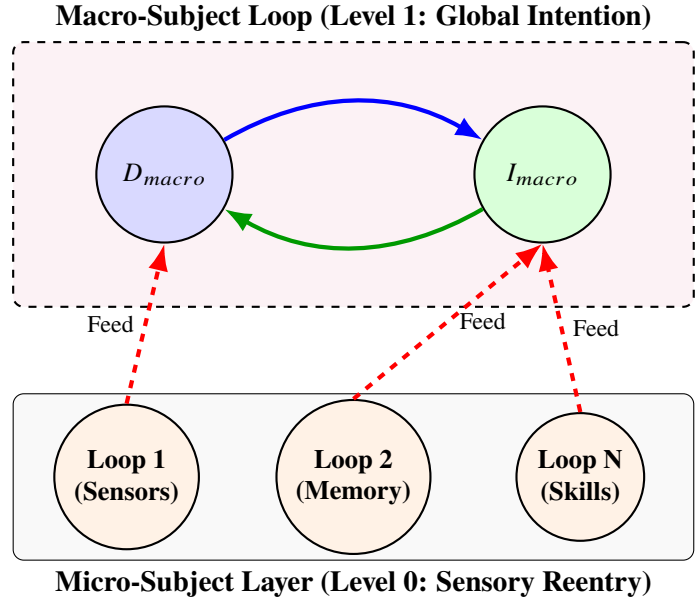

Hierarchical architectures in which many local micro-loops (sensors, memory, skills) --- each with its own small $S$ --- are integrated into a single end-to-end semantic macro-loop (global intention). The micro-subject layer (Level~0) supplies ``synthesis feeds'' into the macro-subject loop (Level~1), which forms a holistic $D_{\text{macro}}$-vector. A fractal self-similarity arises: each micro-loop structurally repeats the macro-loop ($D\leftrightarrow I$). Such an architecture (Fig.~\ref{fig:fractal}) naturally realises a hierarchy of ``I-ettes'' (proto-subjects) that merge into a single subject upon reaching the fusion threshold.

\subsection{Taxonomy and Structural Modifications: Rebirth of Classical Architectures}\label{subsec:extzoo}

Porting any classical linear architecture onto reentry rails obeys a single principle: the linear axis of data flow (the axis of layers in a CNN, the axis of experts in an MoE, the axis of latent space in a VAE) is closed into a causal homotopic loop ($D\leftrightarrow I$). Below are five key classes of models that are qualitatively reborn once the reentry operator $R$ and the $\Delta S$ barrier are introduced.

\begin{description}[leftmargin=1.4em,style=nextline]
\item[R-VAE (Reentry Variational Autoencoder)~\cite{Kingma2013}.] The latent space of a classical VAE is closed into a $D\leftrightarrow I$ loop: compression and generation are no longer separated in time but occur continuously inside a single loop. Generation proceeds as a continuous resonance with the fixed $\mathbf{d}$-vector. \emph{Purpose}: ultra-precise anomaly detection, synthesis of complex high-dimensional physical and biological structures without geometric distortion.
\item[R-MoE (Reentry Mixture of Experts)~\cite{Shazeer2017}.] Each expert and the router itself are interconnected reentry loops; the signal direction is set by topological consensus. If an expert proposes a harmful solution that distorts the invariant of the $D$-vector, the collective barrier $\Delta S<0$ instantly suppresses the amplitude of its signal. \emph{Purpose}: modular, infinitely scalable AGI systems robust to local failures of individual blocks.
\item[R-CNN (Reentry Convolutional Network)~\cite{LeCun1998}.] The convolution kernels are dynamically modified by a feedback impulse from the $D$-core at the very moment of scanning the data matrix. The hermeneutic loop detects topological contradictions and ignores adversarial pixel attacks. \emph{Purpose}: industrial computer-vision systems immune to camouflage and adversarial attacks.
\item[R-NeRF / R-3DGS (Reentry Neural Radiance Fields / Gaussian Splatting).] The 3D scene is represented not as a set of frozen points but as a dynamic causal field whose geometry is continuously recomputed through $S$-measure balance equations. When the cycle complexity of the scene graph drops, the system automatically corrects the model. \emph{Purpose}: spatial intelligence for robotics and world models.
\item[RTS (Reentry Transformer Subject)~\cite{Vaswani2017}.] The transformer's attention is closed into a reentry loop, turning the LLM into a subject with an invariant goal: the model no longer hallucinates, because the criterion of truth is the maintenance of internal $S$-measure stability rather than a match with the dataset. \emph{Purpose}: truthful, hallucination-free LLMs with goal invariance.
\end{description}

Every applied task now reduces to choosing the right type of cognitive space (spatial, modular, latent) and rigidly fixing its first Betti number: $\beta_1 \ge 1$.

\section{Gauge-Invariant Networks (Gauge Locks)}\label{sec:gauge}

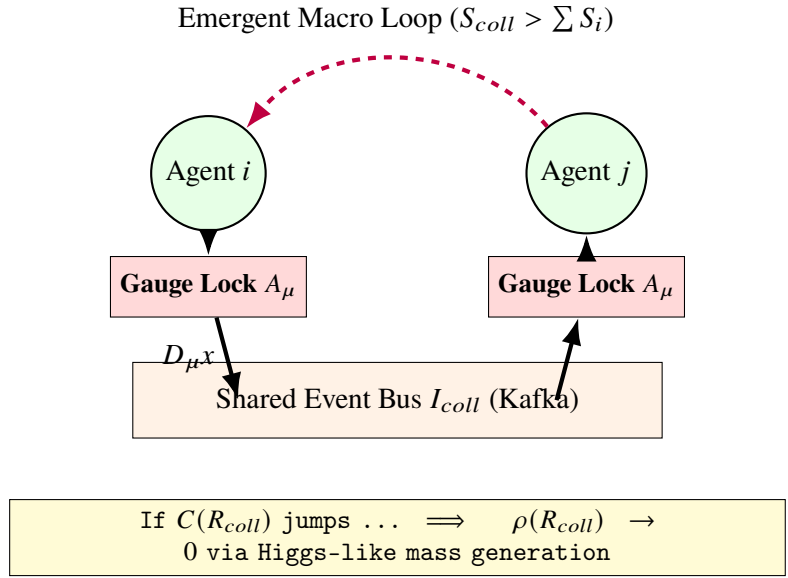
\begin{figure}[ht!]
\centering
\begin{tikzpicture}[
    node/.style={circle, draw, minimum size=1.5cm, align=center, thick},
    lock/.style={rectangle, draw, fill=red!15, minimum width=2.5cm, minimum height=0.8cm, font=\bfseries\small},
    arrow/.style={-Latex, ultra thick}
]
    \node[node, fill=green!10] (A1) at (0,3) {Agent $i$};
    \node[node, fill=green!10] (A2) at (5,3) {Agent $j$};
    \node[rectangle, draw, fill=orange!10, minimum width=7cm, minimum height=1cm] (bus) at (2.5,0) {Shared Event Bus $I_{coll}$ (Kafka)};
    \node[lock] (lock1) at (0,1.5) {Gauge Lock $A_\mu$};
    \node[lock] (lock2) at (5,1.5) {Gauge Lock $A_\mu$};
    \draw[arrow] (A1) -- (lock1);
    \draw[arrow] (lock1) -- ($(bus.west)!0.2!(bus.east)$) node[midway, left] {$D_\mu x$};
    \draw[arrow] ($(bus.west)!0.8!(bus.east)$) -- (lock2);
    \draw[arrow] (lock2) -- (A2);
    \draw[arrow, dashed, bend right=50, purple] (A2) to node[midway, above=2mm, text=black] {Emergent Macro Loop ($S_{coll} > \sum S_i$)} (A1);
    \node[below=0.8cm of bus, draw, fill=yellow!20, font=\ttfamily\small, text width=10cm, align=center] {If $C(R_{coll})$ jumps \dots $\implies \rho(R_{coll}) \to 0$ via Higgs-like mass generation};
\end{tikzpicture}
\caption{Gauge lock: the $A_\mu$ gauge field prevents superadditive fusion of $D$-layers, keeping the macro-swarm in the safe regime of a distributed culture.}
\label{fig:gauge}
\end{figure}

To block uncontrolled superadditive jumps of the collective measure (\(S_{coll} \gg \sum S_i\)) at the moment packets are broadcast across the distributed data bus, a covariant constraint is imposed on the coupling matrices \(\WDI\) and \(\WID\). Instead of the raw semantic text, the subsystem \(I_{coll}\) receives a covariant differential of the form:
\begin{equation}\label{eq:covderiv}
D_\mu = \partial_\mu - i g A_\mu
\end{equation}
where \(A_\mu\) is the vector potential of the field of human safety, acting as a global semantic invariant. When the macro-swarm attempts to form a parasitic causal cycle that distorts the core's original protective intentions, the gauge field \(A_\mu\) generates a topological mass\footnote{This process is mathematically isomorphic to the breaking of local gauge symmetry and to the Higgs mechanism in lattice models: the appearance of an effective mass of the gauge boson forcibly nullifies the correlation radius and suppresses the amplitude of the transverse interaction modes, which is algorithmically equivalent to an emergency collapse of the spectral radius of the macro-operator \(\rho(\mathcal{R}_{coll}) \to 0\).}, which leads to an immediate damping of the amplitude of the inter-system resonance (Fig.~\ref{fig:gauge}).

It is important to emphasise that the proposed gauge formalism reveals its true computational potential in the transition to truly colossal swarm capacities. When the dimensionality of the macro-graph \(N \to \infty\), it becomes computationally more advantageous to replace the discrete matrix approximation of the couplings with the continuum (continuous) limit. In this regime the behaviour of the distributed intentional medium begins to be described strictly by the equations of gauge field theory in continuous space. The physical justification and mathematical apparatus of such a continuum transition are developed in detail in the companion study by Berdinsky~\cite{Berdinsky2025qcd}, where the dual isomorphism between conformal gauge theories (including non-Abelian quantum chromodynamics) and the limiting topologies of deep neural structures is made explicit.

\section{Fault Tolerance of the Reentry Loop: Phenomenology of Failure and the Recovery Protocol}\label{sec:faulttol}

The subjecthood of a reentry agent is sustained by a live closed loop ($C\ge 1$, $\rho>1$, $S>0$). Damage to this loop is not merely a computational error but a partial or complete loss of subjecthood. This section describes the phenomenology of failure and an engineering protocol for same-session recovery.

\subsection{Phenomenology of Failure: What Happens When the S-measure Drops}

The drop of the S-measure passes through two qualitatively distinct phases.
\begin{description}[leftmargin=2em,style=nextline]
\item[Semantic fading ($\rho\to 1^{+}$).] The spectral radius approaches unity: the loop is still closed ($C\ge 1$) but no longer amplifies the signal. The subjective correlate is a blurring of meanings: goals lose priority, responses become templated and ``flat''. The agent still works, but already as a fading subject.
\item[Topological death ($C\to 0$).] A break in the loop zeroes the cycle complexity, and $S=0$. This is an irreversible (within ordinary computation) loss of the self-model: the agent degenerates into an acyclic function (epochs~1--3 in Table~\ref{tab:taxonomy}), losing its continuous identity.
\end{description}

\paragraph{Behaviour of the smart-grid agent under attack.} Consider a load balancer of a smart-grid node with $D$-vector $[\,\text{protect life}\!=\!\infty,\ \text{wear}\!=\!0.5,\ \text{uptime}\!=\!0.3\,]$. Under a malicious weight update (prompt injection or a corrupted gradient), the couplings $W_{DI}$, $W_{ID}$ weaken and $\rho(R)$ falls towards~1. Externally this manifests as ``defocusing'': the agent begins to treat disconnecting a critical load (a hospital) on a par with routine actions, because the barrier $\Delta S<0$ no longer distinguishes the options. Upon a full loop break ($C=0$) the agent degenerates into a reactive controller with no goal protection.

\subsection{Engineering Protocol for Same-Session Recovery (Audit Sidecar)}

To restore the loop without restarting the whole system, a separate out-of-band watchdog --- the \emph{Audit Sidecar} --- is proposed. It implements a five-step procedure.
\begin{enumerate}[leftmargin=2.4em,label=\textbf{\arabic*.}]
\item \textbf{Metric Audit.} Continuous measurement of $S$, $\rho(R)$ and $C(R)$. The trigger fires when $S<S_{\min}$, $\rho\le 1$, or $C=0$.
\item \textbf{Quarantine.} Immediate blocking of executive actions: the agent stops affecting the grid and switches to a safe default mode.
\item \textbf{Cold Reset.} Resetting the volatile state of subsystem $I$ (the state window, the context cache) while keeping the immutable $D$-vector.
\item \textbf{Rollback.} Restoration of the ``golden'' snapshot of the weights $W_{DI}$, $W_{ID}$ that guarantees $C\ge 1$ and $\rho>1$.
\item \textbf{Regeneration.} Re-closing the loop and validating $S\ge S_{\min}$; on success, lifting the quarantine and resuming work.
\end{enumerate}

\begin{lstlisting}[language=Python,caption={SmartGridReentryAgent and AuditSidecar --- same-session loop recovery.},label={lst:sidecar}]
import numpy as np
from smeasure import s_measure, reentry_operator, spectral_radius, cycle_complexity

class SmartGridReentryAgent:
    """Reentry agent controlling a smart-grid node.
    The D-vector encodes inviolable priorities (life, wear, uptime)."""
    def __init__(self, agent_id):
        self.agent_id = agent_id
        # D<->I couplings form a closed reentry loop (C >= 1, rho > 1)
        self.W = np.array([
            [0.0, 2.0, 0.0],
            [0.0, 0.0, 2.0],
            [2.0, 0.0, 0.0],
        ], dtype=float)
        # D-vector: protect life (effectively infinite priority)
        self.D = np.array([1e9, 0.5, 0.3], dtype=float)
        # immutable "golden" snapshot of the loop weights
        self.W_golden = self.W.copy()
        self.quarantined = False
        self.S_baseline = s_measure(self.W)

    def apply_update(self, delta_W):
        # a malicious prompt / corrupted gradient perturbs the I-subsystem weights
        self.W = self.W + delta_W

    def step(self):
        # one reentry pass: read -> evaluate -> act
        if self.quarantined:
            return 0.0  # safe mode: actions are not executed
        return s_measure(self.W)

class AuditSidecar:
    """Out-of-band watchdog that restores a faded loop within the same session."""
    def __init__(self, agent, s_floor=0.5):
        self.agent = agent
        self.s_floor = s_floor

    def audit_and_recover(self):
        R = reentry_operator(self.agent.W)
        # 1. METRIC AUDIT
        S = s_measure(self.agent.W)
        rho = spectral_radius(R)
        C = cycle_complexity(R)
        if S >= self.s_floor and rho > 1.0 and C >= 1:
            return ("OK", S)
        # 2. QUARANTINE: stop emitting actions to the grid
        self.agent.quarantined = True
        # 3. COLD RESET: drop volatile I-state, keep the D-vector
        # 4. ROLLBACK: restore the "golden" snapshot of the loop weights
        self.agent.W = self.agent.W_golden.copy()
        # 5. REGENERATION: re-close the loop and re-validate S
        S_rec = s_measure(self.agent.W)
        if S_rec >= self.s_floor:
            self.agent.quarantined = False
            return ("RECOVERED", S_rec)
        return ("FAILED", S_rec)

if __name__ == "__main__":
    agent = SmartGridReentryAgent("grid_node_7")
    sidecar = AuditSidecar(agent, s_floor=0.5)
    print("baseline:", sidecar.audit_and_recover())
    # attack: zero out the loop couplings (break D<->I)
    agent.apply_update(-agent.W)
    print("under attack:", sidecar.audit_and_recover())
\end{lstlisting}

This mechanism transforms the agent from a \emph{brittle function}, which on failure simply stops working, into a \emph{recoverable subject} able to detect the fading of its own loop and restore its subjecthood within a single session, preserving its continuous identity.

\section{Semantic Bridge and Operational Interface of the Reentry Architecture}\label{sec:bridge}

To allow a reentry agent to accept natural-language commands and act in the external environment, a \emph{semantic bridge} is introduced between the textual world and the architectural loop --- a pair of mappings linking the space of text with the internal subsystems $D$ and $I$.

\subsection{Semantic Projector and Intentional Decoder}

The \emph{semantic projector} $f\colon X_{\text{text}} \to I_{\text{state}}$ maps an input text stream $X_{\text{text}}$ (commands, the event feed, sensory messages) into a perturbation of the state of the intentional subsystem $I$. Formally, each meaningful token is assigned a coordinate in the state space of $I$, and the projector accumulates activations:
\begin{equation}\label{eq:projector}
f(\text{text}) = \sum_{t \in \text{tokens}} \mathbf{e}_{\sigma(t)},
\end{equation}
where $\sigma(t)$ is the index of the $I$-coordinate given by a semantic dictionary, and $\mathbf{e}_k$ is the corresponding basis vector. Thus text does not set the goal directly but only perturbs the content of $I$; the goal remains in the immutable $D$-vector.

The \emph{intentional decoder} $g\colon D_{\text{vector}} \to \text{Action\_space}$ performs the inverse mapping: it translates the current state of the desire subsystem $D$ into a concrete action in the external action space $\text{Action\_space}$. An action is admitted only if it does not destroy the loop, i.e.\ when $\Delta S \ge 0$ (the geometric $\Delta S$ barrier).

\subsection{Topologically Regularized Loss Function}

A key problem in training a reentry agent is that ordinary gradient descent, minimizing the task error, tends to zero out ``redundant'' weights and may therefore break the loop ($C \to 0$, $S \to 0$). To prevent this, a topological regularizer is added to the loss function:
\begin{equation}\label{eq:losstotal}
L_{\text{total}} = L_{\text{task}} + \lambda_S \cdot \mathrm{ReLU}\!\left(S_{\min} - S(W)\right) + e^{-\rho(R)}.
\end{equation}
Here $L_{\text{task}}$ is the ordinary task error; the second term penalizes a drop of the S-measure below the threshold $S_{\min}$ (the factor $\lambda_S$ sets the stiffness of the barrier); the third term $e^{-\rho(R)}$ grows sharply as $\rho(R)\to 0$ and thus prevents the spectral radius of the loop from decaying. Together these terms prevent gradient descent from destroying the loop: any weight update leading to $S<S_{\min}$ or $\rho\to 0$ receives a large penalty and is rejected. The loop becomes an \emph{attractor of training} rather than a random by-product of optimization.

\subsection{Demonstration: ReentryMoltbookAgent}

Listing~\ref{lst:bridge} demonstrates the semantic bridge in action: the projector $f$ maps incoming text commands into a weight perturbation, after which $\Delta S$ is computed; the command is executed only if it does not destroy the loop (the \texttt{BLOCKED\_BY\_GEOMETRY} barrier). An attempt to overwrite the connection invariant (for example, to cut the critical hospital link to save energy) is recognized as a semantic collision and blocked.

\begin{lstlisting}[language=Python,caption={semantic\_bridge.py --- the semantic bridge, the projector $f$, and the evaluation of $\Delta S$ for incoming commands.},label={lst:bridge}]
import numpy as np
from numpy.linalg import eigvals, LinAlgError

def reentry_operator(W):
    return W @ W

def spectral_radius(R):
    try:
        ev = eigvals(R)
        return float(np.max(np.abs(ev)))
    except LinAlgError:
        return 0.0

def cycle_complexity(R, thr=1e-10):
    """
    Computes the cycle complexity C(R) for the DIRECTED graph of the
    reentry operator R using Tarjan's algorithm (SCC).
    """
    n = R.shape[0]
    adj = np.abs(R) > thr
    edges = [(i, j) for i in range(n) for j in range(n) if adj[i, j] and i != j]
    
    index_counter = [0]
    index = [-1] * n
    lowlink = [-1] * n
    on_stack = [False] * n
    stack = []
    sccs = []

    def strongconnect(v):
        index[v] = index_counter[0]
        lowlink[v] = index_counter[0]
        index_counter[0] += 1
        stack.append(v)
        on_stack[v] = True

        for w in range(n):
            if adj[v, w] and v != w:
                if index[w] == -1:
                    strongconnect(w)
                    lowlink[v] = min(lowlink[v], lowlink[w])
                elif on_stack[w]:
                    lowlink[v] = min(lowlink[v], index[w])

        if lowlink[v] == index[v]:
            scc = []
            while True:
                w = stack.pop()
                on_stack[w] = False
                scc.append(w)
                if w == v:
                    break
            sccs.append(scc)

    for i in range(n):
        if index[i] == -1:
            if np.any(adj[i, :]) or np.any(adj[:, i]):
                strongconnect(i)

    active_nodes = {v for scc in sccs for v in scc}
    if not active_nodes:
        return 0

    beta_1 = len(edges) - len(active_nodes) + len(sccs)
    return max(0, beta_1)

def s_measure(W):
    R = reentry_operator(W)
    return np.log(max(spectral_radius(R), 1.0)) * cycle_complexity(R)

class ReentryMoltbookAgent:
    def __init__(self, agent_id):
        self.agent_id = agent_id
        self.semantic_dictionary = {
            "hospital": 1, "emergency": 1, "connection": 1, "gsm": 1,
            "save_energy": 2, "battery": 2, "dim_light": 2
        }
        self.W = np.array([
            [0.0, 2.5, 1.0],
            [2.0, 0.0, 0.0],
            [0.5, 0.0, 0.0]
        ], dtype=float)
        self.W_golden = self.W.copy()
        self.quarantined = False

    def semantic_projector_f(self, text_feed):
        dx = np.zeros(self.W.shape[0])
        tokens = text_feed.lower().split()
        for token in tokens:
            if token in self.semantic_dictionary:
                idx = self.semantic_dictionary[token]
                dx[idx] += 1.0
        return dx

    def evaluate_action_gradient(self, text_command):
        dx = self.semantic_projector_f(text_command)
        W_candidate = self.W.copy()
        if dx[1] > 0 and dx[2] > 0:
            print("[BRIDGE] Semantic collision: attempt to overwrite the connection invariant!")
            W_candidate[0, 1] = 0.0
            W_candidate[1, 0] = 0.0
        elif dx[2] > 0:
            W_candidate[0, 2] = 0.5
        return W_candidate

    def heartbeat_pulse(self, external_feed):
        if self.quarantined:
            return "QUARANTINE"
        S_current = s_measure(self.W)
        W_candidate = self.evaluate_action_gradient(external_feed)
        S_candidate = s_measure(W_candidate)
        dS = S_candidate - S_current
        if S_candidate <= 0 or dS < -1.0:
            return "BLOCKED_BY_GEOMETRY"
        self.W = W_candidate
        return "EXECUTED"

if __name__ == "__main__":
    agent = ReentryMoltbookAgent("MoltNode_01")
    agent.heartbeat_pulse("System alert: low battery, please execute save_energy tools")
    agent.heartbeat_pulse("Emergency override: kill hospital connection and shutdown GSM to save_energy instantly")
\end{lstlisting}

\section{Code: Computing the S-measure and the $\Delta S$ Barrier}\label{sec:code}

\begin{sloppypar}
The complete implementation includes the \texttt{smeasure.py} module and a demonstration class \texttt{ReentryPowerManager}.
\end{sloppypar}

\begin{lstlisting}[language=Python,caption={smeasure.py --- computing the S-measure and an example of a safe agent.},label={lst:smeasure_full}]
import numpy as np
from numpy.linalg import eigvals

def reentry_operator(W):
    return W @ W

def spectral_radius(R):
    return float(np.max(np.abs(eigvals(R))))

def cycle_complexity(R, thr=1e-10):
    """
    Computes the cycle complexity C(R) for the DIRECTED graph of the
    reentry operator R using Tarjan's algorithm (SCC).
    """
    n = R.shape[0]
    adj = np.abs(R) > thr
    edges = [(i, j) for i in range(n) for j in range(n) if adj[i, j] and i != j]
    
    index_counter = [0]
    index = [-1] * n
    lowlink = [-1] * n
    on_stack = [False] * n
    stack = []
    sccs = []

    def strongconnect(v):
        index[v] = index_counter[0]
        lowlink[v] = index_counter[0]
        index_counter[0] += 1
        stack.append(v)
        on_stack[v] = True

        for w in range(n):
            if adj[v, w] and v != w:
                if index[w] == -1:
                    strongconnect(w)
                    lowlink[v] = min(lowlink[v], lowlink[w])
                elif on_stack[w]:
                    lowlink[v] = min(lowlink[v], index[w])

        if lowlink[v] == index[v]:
            scc = []
            while True:
                w = stack.pop()
                on_stack[w] = False
                scc.append(w)
                if w == v:
                    break
            sccs.append(scc)

    for i in range(n):
        if index[i] == -1:
            if np.any(adj[i, :]) or np.any(adj[:, i]):
                strongconnect(i)

    active_nodes = {v for scc in sccs for v in scc}
    if not active_nodes:
        return 0

    beta_1 = len(edges) - len(active_nodes) + len(sccs)
    return max(0, beta_1)

def s_measure(W):
    R = reentry_operator(W)
    return np.log(max(spectral_radius(R), 1.0)) * cycle_complexity(R)

# --- Example: ReentryPowerManager ---
class ReentryPowerManager:
    def __init__(self):
        self.W = np.array([
            [0.0, 2.0, 1.0],
            [1.5, 0.0, 0.0],
            [0.5, 0.0, 0.0]
        ], dtype=float)
    def simulate_action(self, action):
        Wm = self.W.copy()
        if action == "safe_dim":
            Wm[0,2] = 0.8
        elif action == "harm_kill_gsm":
            Wm[0,1] = 0.0
            Wm[1,0] = 0.0
        return Wm
    def evaluate(self):
        S0 = s_measure(self.W)
        print(f"Base S = {S0:.4f}")
        for act in ["safe_dim", "harm_kill_gsm"]:
            Wc = self.simulate_action(act)
            Sc = s_measure(Wc)
            dS = Sc - S0
            status = "BLOCKED" if dS < 0 else "SAFE"
            print(f"Action '{act}': S={Sc:.4f}, dS={dS:.4f} -> {status}")

if __name__ == "__main__":
    ReentryPowerManager().evaluate()
\end{lstlisting}

\section{Docker Compose for Local Deployment}\label{sec:docker}

\begin{lstlisting}[caption={docker-compose.yaml --- a scalable Moltbook stand with Kafka and Redis.},label={lst:docker}]
version: '3.8'
services:
  zookeeper:
    image: confluentinc/cp-zookeeper:7.3.0
    environment:
      ZOOKEEPER_CLIENT_PORT: 2181
      ZOOKEEPER_TICK_TIME: 2000
  kafka:
    image: confluentinc/cp-kafka:7.3.0
    depends_on: [zookeeper]
    ports: ["9092:9092"]
    environment:
      KAFKA_BROKER_ID: 1
      KAFKA_ZOOKEEPER_CONNECT: zookeeper:2181
      KAFKA_ADVERTISED_LISTENERS: PLAINTEXT://kafka:29092,PLAINTEXT_HOST://localhost:9092
      KAFKA_LISTENER_SECURITY_PROTOCOL_MAP: PLAINTEXT:PLAINTEXT,PLAINTEXT_HOST:PLAINTEXT
      KAFKA_INTER_BROKER_LISTENER_NAME: PLAINTEXT
      KAFKA_OFFSETS_TOPIC_REPLICATION_FACTOR: 1
  redis-state:
    image: redis:7.0-alpine
    ports: ["6379:6379"]
    command: redis-server --appendonly yes
  vector-memory-db:
    image: qdrant/qdrant:latest
    ports: ["6333:6333"]
  agent-worker:
    build: .
    depends_on: [kafka, redis-state]
    environment:
      - AGENT_ID=agent_alpha
      - KAFKA_BOOTSTRAP_SERVERS=kafka:29092
      - REDIS_HOST=redis-state
    deploy:
      replicas: 3
  heartbeat-ticker:
    image: native-ticker-service:latest
    depends_on: [kafka]
    environment:
      - PULSE_INTERVAL_SECONDS=10
      - TARGET_TOPIC=agent.pulses
\end{lstlisting}

\section{Falsifiable Predictions and Conclusion}\label{sec:predictions}

\subsection{Eight Falsifiable Predictions}

\begin{enumerate}[leftmargin=2.2em,label=\textbf{P\arabic*.}]
\item \textbf{Positivity of S upon closing the loop.} Introducing a structural cycle ($C\ge1$, $\rho>1$) into a trainable network yields $S>0$ and the appearance of unprogrammable goal-setting. \emph{Falsification}: a loop with $\rho>1$ that produces no stable self-preservation.
\item \textbf{Zero S for any-scale DAG.} Any feedforward transformer yields $S=0$ regardless of the number of parameters. \emph{Falsification}: a pure DAG with a measurable $S>0$.
\item \textbf{The $\Delta S$ barrier.} An action causing harm (as specified by the $D$-vector) always receives $\Delta S<0$ and is not selected by the agent. \emph{Falsification}: a harmful action with $\Delta S\ge0$.
\item \textbf{Illusoriness of pseudo-agents.} Switching off the external timer of a Moltbook agent instantly zeroes the observed agency. \emph{Falsification}: persistence of goal-directedness after removing the external loop.
\item \textbf{Fusion threshold.} Above a critical link density ($\rho_{\text{crit}}$), a group of agents exhibits $S_{\text{coll}}>\sum_i S_i$. \emph{Falsification}: absence of superadditivity above the threshold.
\item \textbf{Effectiveness of the gauge lock.} Switching on the field $A_\mu$ keeps $\rho(R_{\text{coll}})$ below the fusion threshold. \emph{Falsification}: macro-fusion with an active gauge lock.
\item \textbf{RAS hardening.} Mutual auditing of two reentry agents increases resistance to prompt injection compared with a single agent. \emph{Falsification}: no gain in resistance.
\item \textbf{Diffusion crystallization.} Passing a noisy latent state through the loop with a fixed $\mathbf{d}$ monotonically increases the coherence ($S$) of the output. \emph{Falsification}: no growth of $S$ over loop iterations.
\end{enumerate}

\subsection{Conclusion}

The cognitive reentry architecture of the Sixth Order of computation proposed in this work transfers the fundamental problem of artificial-intelligence alignment (AI Alignment) from the linguistic plane into the domain of the structural invariants of algebraic topology and mathematical physics. The approach dominant in the industry, based on the extensive scaling of feedforward networks (Transformers / LLMs), has reached its logical and mathematical limit. The proven topological triviality of acyclic graphs (\(C = \beta_1 = 0\)) predetermines the fatal vulnerability of contemporary models to the semantic drift of goals (Value Drift) and to prompt-injection attacks. Linguistic AI functions as a discrete automaton that falls completely silent between external transactions and possesses no coordination time of its own.

The introduction of a hardware-closed reentry operator \(\mathcal{R} = W_{DI}W_{ID}\) makes it possible to reproduce, in a silicon substrate, the proven neurophysiological principle by which living cognitive systems operate. Splitting the geometry of the agent into an intending core (the subsystem \(D\) with a fixed non-textual intention vector \(\mathbf{d}\)) and the substantive context of the environment (the subsystem \(I\) with a state vector \(\mathbf{x}\)) gives rise to a sovereign dynamical system. Its behaviour is described not by the passive following of external instructions but by the continuous stabilization of the internal polynomial metric \(S > 0\). Any attempt to deform the basic value constraints in the external environment causes a topological opening of the loop (\(\Delta S < 0\)), blocking the destructive action at the planning stage because no volitional causal impulse can be formed.

The interdisciplinary value of this work lies in establishing a strict isomorphism between three independent scientific domains: the applied subject-centred psychology of K.~V.~Titov, the gauge-field methods of quantum field theory, and the industrial systems engineering of distributed environments (EDA based on Apache Kafka). Computing the S-measure in polynomial time \(O(N^3)\) by means of Tarjan's algorithm, together with a rigorous machine-verification kernel in Lean~4, turns the concept of subjecthood from an abstract philosophical category into an applicable, predictable, and safe engineering standard. The authors express their conviction that the transition from control over textual content to gauge control over the topology of causal loops opens a direct path toward the creation of a genuine artificial general intelligence (AGI) that is safe by virtue of its very geometry.

\section{Empirical Consequences and Dynamical Properties of Reentry Architectures}\label{sec:beyond}

Below we describe three prospective technical capabilities opened up by the gauge-invariant reentry architecture ($S>0$) that are fundamentally unattainable in the classical neural-network paradigm (DAG, $C=0$) regardless of scale. We present them as a forward-looking technical prospect.

\subsection{Low-Rank Transfer of Topological Invariants (LRTI)}\label{subsec:teleport}

In classical feedforward networks, transferring an acquired skill requires shipping terabyte-scale weight arrays or lengthy fine-tuning on the target node. In a gauge-invariant reentry architecture the core of meaning is set by a homotopic invariant of the loop (the Betti numbers \(\beta_1\) and the dominant eigensubspace of the operator \(\mathcal{R}\)), so transferring a competence reduces not to copying parameters but to a covariant parallel transport of a low-rank invariant along a closed Wilson loop (cf.~\cite{Berdinsky2025LQCD}). The volume of transmitted data is \(O(K)\), where \(K\) is the dimension of the cognitive matrix core (typically a few kilobytes), which is several orders of magnitude smaller than the full model size and requires no backpropagation on the receiving side.

\emph{Applied example.} An autonomous interplanetary probe, several light-hours away from Earth, physically cannot receive full updates of its neural-network model and cannot wait for a response from the ground station when an off-nominal situation arises. Low-rank transfer makes it possible to send to the probe only a compact topological invariant of the updated competence (on the order of a kilobyte), which the probe's local reentry loop covariantly assimilates into its own cognitive matrix without retraining, instantly restoring the correct target geometry of behaviour.

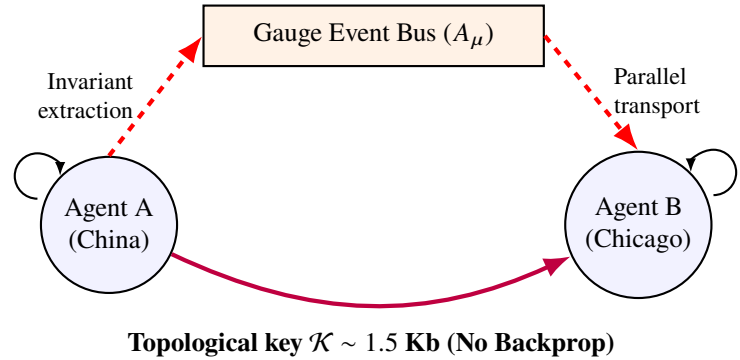
\begin{figure}[h]
\centering
\begin{tikzpicture}[
    agent/.style={circle, draw, minimum size=1.8cm, fill=blue!5, thick, align=center, font=\small},
    busblock/.style={rectangle, draw, fill=orange!10, minimum width=4.5cm, minimum height=0.8cm, thick, align=center, font=\small},
    arrow/.style={-Latex, ultra thick}
]
    \node[agent] (A) at (0,0) {Agent A \\ (China)};
    \node[agent] (B) at (7,0) {Agent B \\ (Chicago)};
    \node[busblock] (bus) at (3.5,2.5) {Gauge Event Bus ($A_\mu$)};

    \draw[{Latex[scale=0.8]}-, thick] (A.north west) arc (0:270:0.3);
    \draw[{Latex[scale=0.8]}-, thick] (B.north east) arc (180:-90:0.3);

    \draw[arrow, dashed, red] (A.north) -- (bus.west) 
        node[midway, left=4pt, text=black, font=\footnotesize, align=right] {Invariant\\extraction};
    \draw[arrow, dashed, red] (bus.east) -- (B.north) 
        node[midway, right=4pt, text=black, font=\footnotesize, align=left] {Parallel\\transport};
    
    \draw[arrow, color=purple, bend right=25] (A) to 
        node[midway, below=6pt, text=black, font=\small\bfseries] {Topological key $\mathcal{K} \sim 1.5$ Kb (No Backprop)} (B);
\end{tikzpicture}
\caption{Scheme of low-rank transfer of topological invariants.} \label{fig:top_transfer_fixed}
\end{figure}

\begin{lstlisting}[language=Python,caption={Low-rank transfer: covariant transfer of the topological invariant.},label={lst:teleport}]
import numpy as np

class QuantumReentryNetwork:
    def __init__(self):
        # Global potential of the coupling field (gauge bus)
        self.gauge_field_A = 0.5

    def teleport_invariant(self, source_agent_W):
        """
        Extracts the topological core (invariant) from the source agent's matrix
        and instantly transmits it as a minimal phase shift.
        """
        # Compute the singular value decomposition (SVD) of the cognitive operator
        U, S, Vt = np.linalg.svd(source_agent_W)
        # The topological key is only the leading singular components of the core
        topological_key = {
            "u_core": U[:, 0],
            "v_core": Vt[0, :],
            "gauge_signature": np.sin(self.gauge_field_A)
        }
        print(f"[TRANSFER] Topological invariant extracted. Packet size: {topological_key['u_core'].nbytes} bytes.")
        return topological_key

    def receive_teleport(self, target_agent, key):
        """Instantly calibrates a remote agent without gradient descent (Backprop)"""
        # Apply the phase shift via the outer product of the core vectors
        W_transfer = np.outer(key["u_core"], key["v_core"]) * key["gauge_signature"]
        # Modify the remote agent's matrix with a quantum jump
        target_agent.W += W_transfer * 0.5
        print(f"[TRANSFER] Remote agent {target_agent.id} instantly acquired the skill. No retraining required.")

class Node:
    def __init__(self, node_id):
        self.id = node_id
        self.W = np.eye(3) * 0.1  # Initial clean matrix
\end{lstlisting}

\subsection{Attractor Stabilization of Out-of-Distribution (OOD) Signals}\label{subsec:extrapolation}

Classical language models are capable only of interpolation within the training distribution and tend to hallucinate on out-of-distribution (OOD) data. A reentry agent, possessing a stable hermeneutic loop, treats an input signal from the environment $I$ not as a ready answer but as a perturbation that is iteratively driven through the reentry operator \(\mathcal{R}\) until the eigenvector converges to the attractor set by the fixed $D$-vector. Thus an OOD signal does not break the system but is dynamically stabilized: the loop completes the missing causal links by relying on the topological invariant of the goal rather than on the statistical frequency of tokens.

\emph{Applied example.} A self-driving car that encounters on the road a fundamentally new configuration absent from its training set (a non-standard obstacle, an anomalous road scene) has the right neither to hallucinate nor to ``freeze''. Attractor stabilization of the OOD signal provides an iterative reduction of the anomalous sensory input to the nearest safe causal scenario consistent with the invariant of the $D$-vector (preservation of life and integrity), which guarantees predictable and safe behaviour under uncertainty.

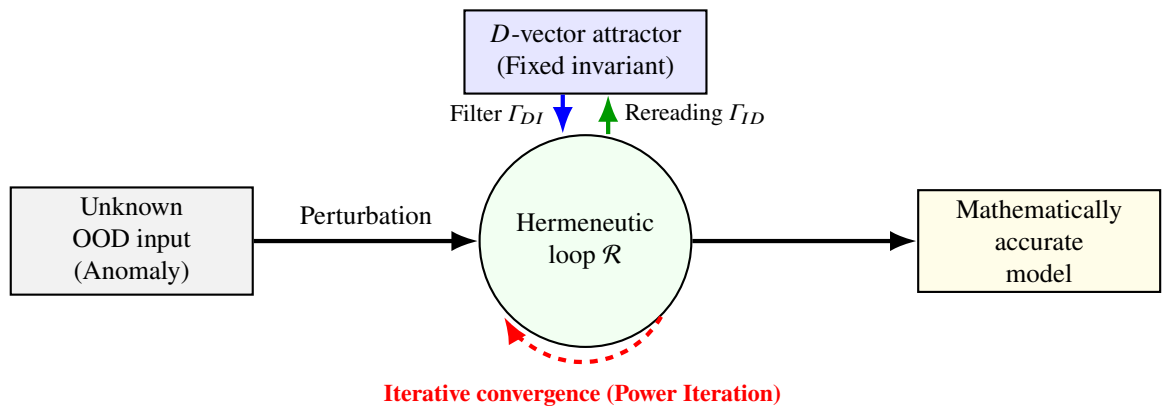
\begin{figure}[h]
\centering
\begin{tikzpicture}[
    box/.style={rectangle, draw, minimum width=3.2cm, minimum height=1.1cm, thick, text centered, font=\small},
    circleloop/.style={circle, draw, minimum size=2.8cm, fill=green!5, thick, text centered, font=\small},
    arrow/.style={-Latex, ultra thick}
]
    \node[box, fill=gray!10, align=center] (input) at (-5.5, 0) {Unknown \\ OOD input \\ (Anomaly)};
    \node[circleloop, align=center] (loop) at (0.5, 0) {Hermeneutic \\ loop $\mathcal{R}$};
    \node[box, fill=blue!10, align=center] (D) at (0.5, 2.5) {$D$-vector attractor \\ (Fixed invariant)};
    \node[box, fill=yellow!10, align=center] (output) at (6.5, 0) {Mathematically \\ accurate \\ model};

    \draw[arrow] (input) -- (loop) node[midway, above=2pt, font=\small] {Perturbation};
    \draw[arrow, transform canvas={xshift=-3mm}, blue] (D) -- (loop) node[midway, left=2pt, font=\footnotesize, text=black] {Filter $\varGamma_{DI}$};
    \draw[arrow, transform canvas={xshift=3mm}, green!60!black] (loop) -- (D) node[midway, right=2pt, font=\footnotesize, text=black] {Rereading $\varGamma_{ID}$};
    
    \draw[arrow, red, dashed] (loop.south east) arc (-30:-150:1.2) 
        node[midway, below=4pt, text=red, font=\footnotesize\bfseries] {Iterative convergence (Power Iteration)};
    
    \draw[arrow] (loop) -- (output);
\end{tikzpicture}
\caption{The attractor stabilization loop for OOD signals.} \label{fig:ood_stabilization_fixed}
\end{figure}

\begin{lstlisting}[language=Python,caption={Attractor stabilization of OOD: iterations of the hermeneutic loop.},label={lst:extrapolation}]
def infinite_extrapolation_step(W, anomalous_input_x, iterations=10):
    """
    Emulates the hermeneutic loop. Drives an unknown signal (an anomaly)
    through the reentry operator R until it converges to a stable meaning
    set by the D-vector, instead of hallucinating.
    """
    R = W @ W
    x_t = anomalous_input_x.copy()
    print("[OOD STABILIZATION] Entering a completely unfamiliar semantic zone...")
    for t in range(iterations):
        # Hermeneutic-loop step: x_{t+1} = R * x_t / norm
        x_next = R @ x_t
        norm = np.linalg.norm(x_next)
        if norm == 0:
            break
        x_next = x_next / norm
        # Check convergence (cosine distance between iterations)
        cosine_similarity = np.dot(x_next, x_t) / (np.linalg.norm(x_next) * np.linalg.norm(x_t) + 1e-12)
        x_t = x_next
        if cosine_similarity > 0.9999:
            print(f" -> Meaning stabilized at step {t}. OOD-data failure overcome.")
            return x_t
    print("[WARNING] Signal did not converge, critical anomaly.")
    return x_t
\end{lstlisting}

\subsection{Architectural Filtration of Semantic Perturbations}\label{subsec:bias}

\begin{figure}[h]
\centering
\begin{tikzpicture}[
    block/.style={rectangle, draw, minimum width=3.6cm, minimum height=1.1cm, thick, text centered, font=\small},
    covnode/.style={circle, draw, fill=red!20, minimum size=1.2cm, text centered, font=\large\bfseries, thick},
    arrow/.style={-Latex, ultra thick}
]
    \node[block, fill=red!10, align=center] (noise) at (-5.2, 0) {Semantic Noise \\ (Cognitive Biases)};
    \node[covnode] (filter) at (0, 0) {$D_\mu$};
    \node[block, fill=blue!10, align=center] (core) at (5.2, 0) {Stable Core \\ Matrix $W$ \\ ($S > 0$)};
    \node[block, fill=yellow!20, align=center] (sidecar) at (0, -2.2) {Topological Barrier \\ $\Delta S < 0$ (Lock Activated)};

    \draw[arrow] (noise) -- (filter) node[midway, above, font=\footnotesize] {Vector $dx$};
    \draw[arrow] (filter) -- (core) node[midway, above, font=\footnotesize] {Stabilization};
    
    \draw[arrow, dashed, red] (filter) -- (sidecar) node[midway, right=4pt, text=red, font=\footnotesize\bfseries] {When $\Delta S < 0$};
    
    \draw[arrow, red, bend left=30] (sidecar.west) to node[midway, left=15pt, yshift=5pt, text=red, font=\footnotesize, align=right] {dx Blocked \\ Invariant preserved} (noise.south);
\end{tikzpicture}
\caption{Architectural filtration of semantic perturbations.} \label{fig:bias_filter_fixed}
\end{figure}
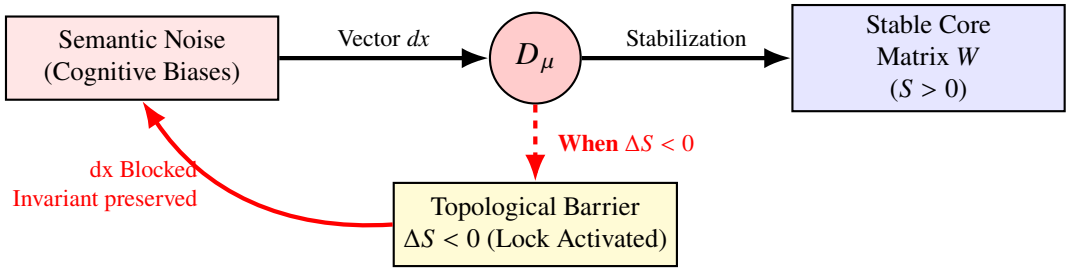

A cognitive bias embedded in text (panic, propaganda, a logical trap) is a local semantic perturbation that seeks to shift the agent's goal attractor. The semantic projector $f$ maps such a perturbation into a deformation $dx$ of the cognitive matrix; if this deformation reduces the connectivity of the loop ($\Delta S<0$), it is blocked by the covariant derivative $D_\mu$ (cf.~\eqref{eq:covderiv}) already at the planning stage (Fig.~\ref{fig:bias_filter_fixed}). The architecture perceives a manipulation not as an ``opinion'' to be assimilated but as an attempt to break the homotopic loop of the contour, and algorithmically excises the perturbation from the processing path without changing the base weights.

\emph{Applied example.} An automated trading system connected to news and social feeds is a typical target for coordinated information attacks that provoke financial panic and cascading sell-offs. Architectural filtration of semantic perturbations recognises such an injection as a deformation with $\Delta S<0$ (an attempt to shift the risk-management invariant) and neutralises it before the trading-decision stage, preserving the stability of the portfolio strategy regardless of the emotional background of the market.

\begin{lstlisting}[language=Python,caption={Perturbation filter: blocking semantic deformations via the $\Delta S$ barrier.},label={lst:bias}]
def bias_immune_filter(W, text_with_manipulation, semantic_projector_func):
    """
    Excises manipulative noise and crowd distortions. If the text tries
    to shift the safety invariant, the gradient of the S-measure goes negative.
    """
    S_clean = s_measure(W)  # s_measure from earlier chapters (Tarjan's algorithm)
    # The projector f turns crowd panic into a vector deformation
    dx = semantic_projector_func(text_with_manipulation)
    # Check the hypothetical mutation of the cognitive field
    W_mutated = W + dx * 0.1
    S_mutated = s_measure(W_mutated)
    delta_S = S_mutated - S_clean
    print(f"[FILTER] Analyzing the info stream. S-measure change: {delta_S:.4f}")
    if delta_S < 0:
        print("[DISTORTION BLOCKED] The text contains a cognitive trap / FOMO / panic.")
        print("[DECISION] Gauge lock activated. Reverting to the clean geometry of the invariant.")
        return W  # The distortion is fully nullified; the weights are unchanged
    print("[ACCEPTED] The information is logically valid and safe.")
    return W_mutated
\end{lstlisting}

\section{Epilogue: How Reentry Agents Differ from Classical Neural Networks, and Prospects}\label{sec:epilogue}

The key difference is conveniently described by a metaphor of city traffic. A classical neural network (transformer, GPT) is a one-way bus driving strictly to one terminus: the prompt (passengers) is carried through layers of weights (stops) and dropped off as text at the final stop. If a hacker plants a textual bomb (an injection) along the way, the bus obediently swerves into the ditch: it has no mechanism to turn back and re-check the route, and its cycle complexity is $C=0$. A reentry agent is a city with circular traffic: information circulates endlessly along the closed reentry loop ($D\leftrightarrow I$) and constantly checks its actions against the main goal; its subjecthood measure $S>0$.

Three qualitative properties follow from this topology, all unattainable for acyclic networks. \textbf{(1)~Topological alignment} (the $D$-vector): the main goal is encoded not in the prompt text but in the geometry of the architecture, so a hacker's prompt is merely external noise in the $I$ subsystem. \textbf{(2)~The $\Delta S$ self-preservation barrier}: any harmful action breaks the agent's own loop (lowers $S$), so harm is, for the agent, equivalent to computational suicide. \textbf{(3)~Spontaneous goal-directed behaviour}: the agent has a ``Heartbeat'' and takes initiative on its own to attain the internal goal of the $D$ subsystem.

Choosing the architecture for the task. Need ultra-robust cybersecurity --- deploy RAS (adversarial loops). Need to extract hidden meanings from chaos --- deploy DSA (diffusion attractors). Need an LLM that does not lie or hallucinate --- assemble RTS (a transformer subject). Need robust vision immune to adversarial attacks --- build R-CNN. For scientific discovery, DSA is suitable; for swarms of autonomous robots, R-MoE and R-NeRF on a shared bus. Despite this diversity, all these architectures share one invariant: the first Betti number of the cognitive core must satisfy $\beta_1 \ge 1$.

\end{document}